\documentclass[10pt,twocolumn,letterpaper]{article}

\usepackage{cvpr}              % To produce the CAMERA-READY version
\usepackage{graphicx}
\usepackage{amsmath}
\usepackage{amssymb}
\usepackage{booktabs}
\usepackage{bm}
\usepackage{xr}
\usepackage{amsthm}
\usepackage{booktabs}
\usepackage{algorithm}
\usepackage{algorithmic}
\usepackage{bbding}
\usepackage{pifont}
\usepackage[switch]{lineno}   %% <- no mathlines option
\usepackage{amsmath}  %% <- after lineno
\usepackage{etoolbox} %% <- for \cspreto, \csappto
\newcommand{\cmark}{\ding{51}}
\newcommand{\xmark}{\ding{55}}

\usepackage[pagebackref,breaklinks,colorlinks]{hyperref}

\usepackage[capitalize]{cleveref}
\crefname{section}{Sec.}{Secs.}
\Crefname{section}{Section}{Sections}
\Crefname{table}{Table}{Tables}
\crefname{table}{Tab.}{Tabs.}

\def\cvprPaperID{*****} % *** Enter the CVPR Paper ID here
\def\confName{CVPR}
\def\confYear{2022}

\begin{document}

\title{Parallel Pre-trained Transformers (PPT) \\ for Synthetic Data-based Instance Segmentation}

\author{ Ming Li\footnotemark[1] 
\quad \quad \quad Jie Wu\footnotemark[1] 
\quad \quad \quad Jinhang Cai
\quad \quad \quad Jie Qin
\quad \quad \quad Yuxi Ren \\
 Xuefeng Xiao 
 \quad \quad \quad Min Zheng
 \quad \quad \quad Rui Wang
  \quad \quad \quad Xin Pan 
 \\
ByteDance Inc.\\
% {\tt\small \{renyuxi.20190622, wujie.10, xiaoxuefeng.ailab, yangjianchao\}@bytedance.com}
}

\maketitle
\footnotetext [1]{Equal Contribution. Corresponding author is Jie Wu.}

%%%%%%%%% ABSTRACT
\begin{abstract}
Recently, Synthetic data-based Instance Segmentation has become an exceedingly favorable optimization paradigm since it leverages simulation rendering and physics to generate high-quality image-annotation pairs.
In this paper, we propose a \textit{\textbf{Parallel Pre-trained Transformers (PPT)}} framework to accomplish the synthetic data-based Instance Segmentation task.
Specifically, we leverage the off-the-shelf pre-trained vision Transformers to alleviate the gap between natural and synthetic data, which helps to provide good generalization in the downstream synthetic data scene with few samples.
Swin-B-based CBNet V2, Swin-L-based CBNet V2 and Swin-L-based Uniformer are employed for parallel feature learning, and the results of these three models are fused by pixel-level Non-maximum Suppression (NMS) algorithm to obtain more robust results. 
The experimental results reveal that PPT ranks first in the CVPR2022 AVA Accessibility Vision and Autonomy Challenge, with a 65.155\% AP.

\end{abstract}

%%%%%%%%% BODY TEXT
\section{Introduction}
\label{sec:intro}

Instance segmentation is a fundamental and crucial task due to extensive vision applications such as scene understanding and autonomous driving. However, this task is still highly dependent on adequate granular pixel-level annotations, which requires a considerable amount of manual effort.
To alleviate such expensive and unwieldy annotations, some researchers attempt to address this task via synthetic data, where a series of simulation rendering and physics are designed to provide high-quality image-annotation pairs.
To provide vision-based benchmarks relevant to accessibility, "CVPR2022 AVA Accessibility Vision and Autonomy Challenge" is set up to spark the interest of  AI researchers working on more robust visual reasoning models for accessibility and boost the performance of synthetic data-based instance segmentation~\cite{xworld}.
The challenge involves a synthetic instance segmentation benchmark incorporating use-cases of autonomous systems interacting with pedestrians with disabilities, which contains challenging accessibility-related person and object categories, such as `cane' and `wheelchair.'
Compared with traditional instance segmentation, this challenge confronts three additional difficulties:
(i) domain generalization problem between synthetic and natural data.
(ii) long-tailed and few sample issues.
(iii) uncertain problem of segmentation in the edge.

\begin{figure*}[t]
  \centering
  \includegraphics[width=1.0\textwidth]{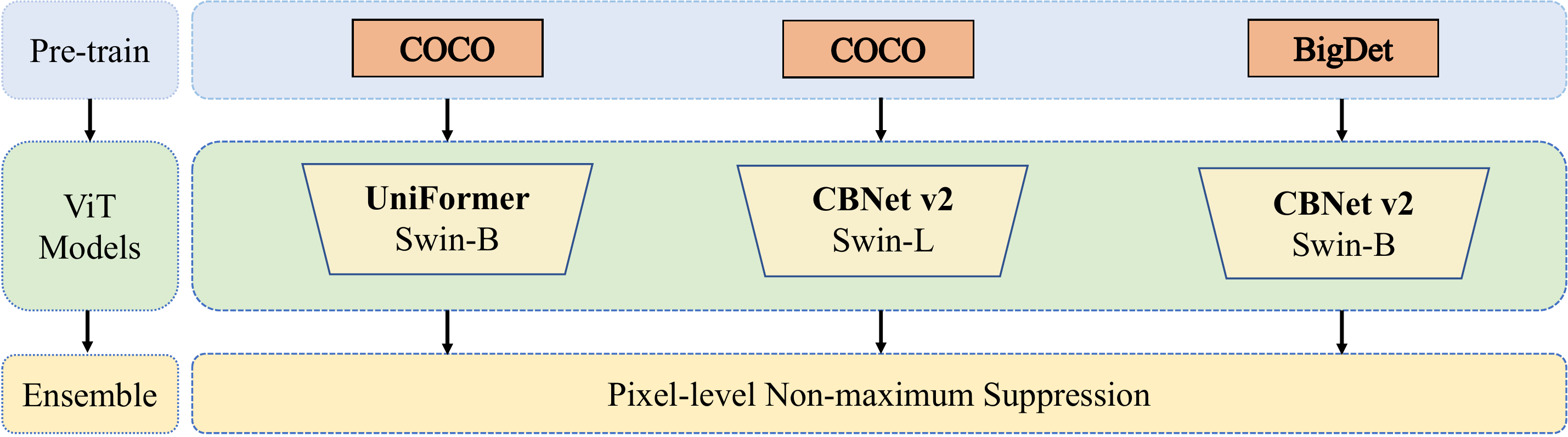}
  \caption{The pipeline of our Parallel Pre-trained Transformers (PPT).}
    \label{fig:pipeline}
\end{figure*}

This paper proposes a \textit{\textbf{Parallel Pre-trained Transformers (PPT)}}  framework to accomplish this challenge.
Some studies have revealed that the model pre-trained with large-scale datasets can generalize well in diverse downstream scenarios.
Hence we use the off-the-shelf pre-trained model (from COCO~\cite{coco} and BigDetection~\cite{bigdet}) to alleviate the gap between natural and synthetic data so that it can be quickly generalized in the downstream synthetic data scene with few samples.
We employ Swin-B-based CBNet V2, Swin-L-based CBNet V2~\cite{cbnetv2, swin}, and Swin-L-based Uniformer~\cite{uniformer} for parallel feature learning, and the results of these thress models are fused by pixel-level NMS~\cite{solo} to obtain more robust results. In the training process, we design a balance copy-and-paste data augmentation approach to alleviate the problem of few samples and long-tail distribution. 
Experimental results on “CVPR2022 AVA Accessibility Vision and Autonomy Challenge” show that our proposed PPT rank first in the competition.

%-------------------------------------------------------------------------
\section{Methodology}
In this section, we first illustrate the overview of the proposed PPT framework in Figure \ref{fig:pipeline}. 
Then the pre-trained Transformers and data augmentation are presented in Sec.~\ref{sec: pt} and Sec.~\ref{sec: da}, respectively. The pixel-level non-maximum suppression is introduced Sec.~\ref{sec: fusion}.  

\subsection{Pre-trained Transformer}
% COCO and BigDet
\label{sec: pt}
Pre-trained models have demonstrated their strong generalization performance on diverse tasks, i.e., the models trained on large-scale datasets can quickly and well transfer to other datasets~\cite{moco,mae}. Since Vision Transformer (ViT) does not have the inductive bias introduced by CNN, it can capture more generalized and robust global concepts on large-scale datasets~\cite{vit}. Therefore, we choose to fine-tune a series of competitive ViT-based detection models, which are pre-trained on large-scale object detection or instance segmentation datasets like BigDetection~\cite{bigdet} and COCO~\cite{coco}.

UniFormer~\cite{uniformer} integrates the advantages of CNN and self-attention in a concise transformer format, which explores local and global token affinity in shallow and deep layers, respectively. CBNet v2 ~\cite{cbnetv2} groups multiple identical backbones through composite connections to construct high-performance detectors. BigDetection~\cite{bigdet} is a larger dataset for improved detector pre-training by simply combining the training data from existing datasets (LVIS~\cite{lvis}, OpenImages~\cite{openimages} and Objects365~\cite{objects365}) with carefully designed principles. The pre-trained CBNet v2 on BigDetection has shown significant generalization ability~\cite{bigdet}.

\subsection{Data Augmentation}
% simple copy-paste
\label{sec: da}
Data augmentation is essential for many vision tasks. Multi-scale training is a common and valuable data augmentation used for object detection and instance segmentation. Although it has a good performance improvement for datasets with multi-scale images like COCO~\cite{coco}, it does not improve for the current challenge dataset, which we assume is because all images are of the same size.

Simple copy-paste~\cite{scp} technology provides impressive gains for the instance segmentation model by pasting objects randomly, especially for datasets under long-tailed distribution. However, we argue that this simple approach equally increases the samples for all classes but can not provide additional supervision signals for the categories with few samples. Therefore, we propose the balanced-copy-paste data augmentation approach. Balanced-copy-paste adaptively samples the objects according to their effective number of objects~\cite{effective_num}, which further alleviates the long-tail effect and improves the overall performance.

\subsection{Pixel-level Non-maximum Suppression}
% TTA, Mask NMS
\label{sec: fusion}
We use Test-Time Augmentation (TTA) to obtain more robust prediction results. Specifically, we resize the test image to different scales and fuse the predictions from multiple scales to get the final result, which can be seen as the fusion of a single image under multiple data augmentation.

Model ensemble is an effective way to improve final performance by integrating the outputs of different models to obtain more robust and complete prediction results. After training all single models and obtaining mask predictions with TTA, we simply ensemble all results by pixel-level NMS~\cite{solo} with Gaussian kernel of 2.0.

\section{Experiments} 
\subsection{Experimental Settings.}
\paragraph{Models}
In this paper, we use Swin-Base-based CBNet V2, Swin-Large-based CBNet V2 and Swin-Base-based Uniformer. 
We leverage Big-Detection pre-trained weights to optimize Swin-Base-based CBNet V2 and employ COCO pre-trained weights for other models. These pre-trained weights are taken from their official release. We also provide smaller detection models for analysis, such as Mask RCNN~\cite{maskrcnn} and DetectoRS~\cite{detectors}.

\paragraph{Datasets.}
We adopt the challenge data~\cite{xworld} for training and testing. It should be noted that this dataset is exposed in long-tailed data distribution, and the number of instances of normal pedestrians and vehicles is much higher than that of accessibility-related person and object categories.

\paragraph{Implementation Details.}
We combine the training and validation sets to train our model with the input image size $1920 \times 1080$. Padding is used to make sure each side can be divided by 32. We adopt simple-copy-paste~\cite{scp} data augmentation with a 0.5  probability horizontal flip to relieve the long-tailed distribution issue.
Training hyper-parameters such as learning rate and weight decay mostly follow the setting from the COCO dataset~\cite{coco}. Test-Time Augmentation (TTA) is used to obtain more competitive results.

\begin{table}[!]
\resizebox{1.0\linewidth}{!}{
\begin{tabular}{c|c|c|c|c}
\toprule
Method        & Backbone & Pretrain      & Epochs & mAP (Val)  \\
\toprule
Mask RCNN$^*$ & R50      & Rand          & 12     & 31.9 \\
Mask RCNN$^*$ & R50      & ImageNet      & 12     & 42.0 \\
Mask RCNN$^*$ & R50      & COCO          & 12     & 44.3 \\
\midrule
DetectoRS     & R101     & COCO          & 12     & 54.7 \\
DetectoRS     & R101     & COCO          & 36     & 55.2 \\
\midrule
UniFormer     & Swin-B   & COCO          & 36     & 59.0 \\
HTC           & Swinv2-L & ImageNet      & 36     & 59.2 \\
CBNet v2      & Swin-B   & BigDet        & 36     & 60.6 \\
CBNet v2      & Swin-L   & COCO          & 36     & 60.8 \\
\bottomrule
\end{tabular}
}
\caption{Instance segmentation mAP on validation dataset for different methods. Better pre-trained weights, stronger augmentation, longer training epochs and larger models can improve the final result. $*$ denotes we use simple augmentation, and stronger augmentation and the original image size is used by default.}
\label{tab:models}
\end{table}

\subsection{Experimental Results}
\paragraph{Better Pre-trained Weights}
We extensively evaluate the effectiveness of different models and pre-trained weights in Table~\ref{tab:models}. We use Mask RCNN with ResNet-50~\cite{resnet} to evaluate different pre-training weights. When the model weights are randomly initialized, the final mAP is $31.9\%$ on the validation dataset. ImageNet pre-train (backbone only) and COCO pre-train (entire model) significantly boost the performance, which improves $10.1\%$ and $12.3\%$ mask mAP, respectively. For CBNet v2, the result of BigDet pre-training with swin-B backbone is close to the COCO pre-training with swin-L backbone, which further indicates that better pre-trained leads to better performance improvement.

\paragraph{Stronger Models}
In addition to Mask RCNN~\cite{maskrcnn}, we also evaluate other stronger models that work better for COCO instance segmentation\cite{coco}, including DetectoRS, UniFormer, HTC, and CBNet v2 ~\cite{detectors,uniformer,htc,cbnetv2} in Table~\ref{tab:models}. The mAP of these four models on COCO is incremental, and their fine-tune performance on the challenge dataset is also incremental, which means that the stronger pre-trained models have better fine-tune results.

\paragraph{Improved Training Strategies}
We evaluate different training strategies in Table~\ref{tab:train_strageties}, including different input resolutions, whether to use multi-scale training, whether to use SCP and different mask loss weights.  Multi-scale training does not bring performance gains, it may be because all images are of $1920\times1080$ resolution and the fixed size is enough.
Copy-paste augmentation and increasing the weight of mask loss can further improve the final result, \emph{i.e.,} from baseline $44.3\%$ mAP to $51.7\%$ mAP for Mask RCNN. In order to fully unleash the power of the large model, we also include the validation set into the training process, adjust the test hyper-parameters and introduce TTA. These strategies improve the UniFormer and CBNet v2 by $2.4\%$ mAP and $2.5\%$ mAP, respectively.

\begin{table}[t!]
\resizebox{1.0\linewidth}{!}{
\begin{tabular}{c|c|c|c|c}
\toprule
Method      & Training Data & TTA      & Hyper-parm     & mAP (Test)  \\
\toprule
UniFormer   & Train         & \xmark   & default        & 60.2 \\
UniFormer   & Train         & \cmark   & default        & 60.8 \\
UniFormer   & Train         & \cmark   & optimal        & 61.4 \\
UniFormer   & Train + Val   & \cmark   & optimal        & 62.6 \\
\midrule
CBNet v2    & Train         & \xmark   & default        & 61.7 \\
CBNet v2    & Train         & \cmark   & default        & 62.1 \\
CBNet v2    & Train         & \cmark   & optimal        & 62.6 \\
CBNet v2    & Train + Val   & \cmark   & optimal        & 64.2 \\
\bottomrule
\end{tabular}
}
\caption{Results on the test dataset. TTA denotes we apply test-time augmentation during testing. Hyper-parm denotes the test hyper-parameters, default means standard setting for each method, optimal denotes the optimal test hyper-parameters on the validation dataset. All the models are trained with mask loss weight 1.0.}
\label{tab:test_strageties}
\end{table}

\paragraph{Improved Test Strategies}
We evaluate different test strategies for UniFormer and CBNet v2 in Table~\ref{tab:test_strageties}. TTA denotes test-time augmentation and enhances  $0.6\%$ and $0.4\%$ mAP for UniFormer and CBNet v2, respectively. We also search for different test hyper-parameters (such as the type and threshold of NMS) on CBNet v2, and apply them to other models. UniFormer and CBNet v2 achieve the mAP performance of $61.4\%$ and $62.6\%$ on the test dataset with the best hyper-parameters, respectively.
 
\begin{table}[!]
\resizebox{1.0\linewidth}{!}{
\begin{tabular}{c|c|c|c|c}
\toprule
Resolution       & MS-Train & CP      & $\lambda_{mask}$ & mAP (Val) \\
\toprule
$1333\times800$  & \xmark   & None    & 1.0          & 44.3 \\
$1333\times800$  & \cmark   & None    & 1.0          & 45.3 \\
\midrule
$1920\times1080$ & \xmark   & None    & 1.0          & 50.3 \\
$1920\times1080$ & \cmark   & None    & 1.0          & 50.3 \\
\midrule
$1920\times1080$ & \xmark   & Simple  & 1.0          & 50.9 \\
$1920\times1080$ & \xmark   & Simple   & 1.2          & 51.4 \\
$1920\times1080$ & \xmark   & Simple   & 1.6          & 51.0 \\
$1920\times1080$ & \xmark   & Simple   & 2.0          & 51.4 \\
\midrule
$1920\times1080$ & \xmark   & Balance  & 2.0          & 52.1 \\
\bottomrule
\end{tabular}
}
\caption{Results with different training strategies on the validation dataset. All the experiments are evaluated with Mask RCNN (R50) training with standard 1x strategy, and validation is not used during training. MS-Train denotes multi-scale training, CP denotes copy-paste, and $\lambda_{mask}$ is the weight of mask loss.}
\label{tab:train_strageties}
\end{table}

\paragraph{Model Ensemble}
We combine the mask results of all models after post-processing and obtain the final result by mask NMS, which can be seen as a parallel model ensemble. The ensemble result with Gaussian kernel is $0.9\%$ MAP higher than the best single model, as illustrated in Table~\ref{tab:ensemble_val}. Finally, we ensemble the best models and obtain the mask mAP of 65.2\% on the test dataset, as shown in Table~\ref{tab:ensemble_test}.

\begin{table}[t!]
\resizebox{1.0\linewidth}{!}{
\begin{tabular}{c|c|c|c|c}
\toprule
Method     & Backbone & Pretrain  & Data  & mAP (Val) \\
\toprule
UniFormer  & Swin-B   & COCO      & Train & 60.0      \\
CBNet v2   & Swin-L   & COCO      & Train & 61.8      \\
CBNet v2   & Swin-B   & BigDet    & Train & 61.6      \\
\toprule
Ensemble (linear)     & -        & -         & Train & 62.6      \\
Ensemble (Gaussian)   & -        & -         & Train & 62.7      \\
\bottomrule
\end{tabular}
}
\caption{Ensemble ablation study on the validation dataset. In order to make the ensemble more reasonable and effective, the models here only use the training dataset. We use the optimal ensemble config for the final ensemble on the test dataset.}
\label{tab:ensemble_val}
\end{table}

\begin{table}[t!]
\resizebox{1.0\linewidth}{!}{
\begin{tabular}{c|c|c|c|c}
\toprule
Method     & Backbone & Pretrain  & Data  & mAP (Test) \\
\midrule
UniFormer  & Swin-B   & COCO         & Train+Val   & 63.3      \\
\midrule
CBNet v2   & Swin-L   & COCO         & Train+Val   & 64.7      \\
CBNet v2   & Swin-B   & BigDet         & Train+Val   & 64.2      \\
\midrule
Ensemble   & -        & -           & Train+Val   & 65.2      \\
\bottomrule
\end{tabular}
}
\caption{Ensemble results with Gaussian kernel on the test dataset. CBNet v2 with backbone Swin-L is pre-trained on COCO and CBNet v2 with backbone Swin-B is pretarined on BigDet. All the results are evaluated with TTA and optimal searched test config.}
\label{tab:ensemble_test}
\end{table}

%------------------------------------------------------------------------
\section{Conclusion}
In this paper, we propose a \textit{Parallel Pre-trained Transformers (PPT)} framework to accomplish the synthetic data-based Instance Segmentation task.
Swin-B-based CBNet V2, Swin-L-based CBNet V2 and Swin-L-based Uniformer are employed for parallel feature learning, and the results of these four models are fused by mask NMS to obtain more robust results. 
Experimental results on “CVPR2022 AVA Accessibility Vision and Autonomy Challenge” demonstrate that PPT rank first in this competition.

%%%%%%%%% REFERENCES
{\small
\bibliographystyle{ieee_fullname}
\bibliography{egbib}
}

\end{document}